\documentclass{IEEEtran}
\usepackage{amsmath,amsthm,amssymb,amsfonts}
\usepackage{hyperref}
\usepackage{amssymb}

\usepackage[labelfont=bf]{caption}
\usepackage[justification=centering]{caption}
\usepackage{amsmath}
\usepackage{color, soul}
\usepackage{cite}
\usepackage{longtable}
\usepackage{hyperref}
\usepackage{array}
\usepackage{float}
\usepackage{titlesec}
\usepackage{array,multirow,graphicx}

\usepackage{tikz}
\usepackage{breqn}
\usepackage{enumitem}
\usepackage{multirow}
\usepackage{algorithm}
\usepackage{balance}
\usepackage{eurosym}
\usepackage{tabularx}
\usepackage{caption}
\usepackage{picture}
\usepackage{lscape}
\usepackage{xcolor}
\usepackage{setspace}
\usepackage{pdflscape}
\usepackage{makecell}
\usepackage{supertabular,booktabs}

\usepackage{subfigure}

\usepackage{array}
\usepackage{calc}

\title{A Guide to Employ Hyperspectral Imaging for Assessing Wheat Quality at Different Stages of Supply Chain in Australia: A Review}

\author{Priyabrata Karmakar, Shyh Wei Teng. Manzur Murshed, Paul Pang, Cuong Van Bui
	\thanks{Priyabrata Karmakar, Shyh Wei Teng. Manzur Murshed, Paul Pang, Cuong Van Bui are with the Institute of Innovation, Science and Sustainability, Federation University Australia (email:{p.karmakar,shyh.wei.teng, manzur.murshed, p.pang, v.bui}@federation.edu.au).}
	\thanks{
		This work was supported by Australian Government Traceability Grants Program Round 2. }}

\begin{document}
\maketitle
	
	\begin{abstract}
Wheat is one of the major staple crops across the globe. Therefore, it is mandatory to measure, maintain and improve the wheat quality for human consumption. Traditional wheat quality measurement methods are mostly invasive, destructive and limited to small samples of wheat. In a typical supply chain of wheat, there are many receival points where bulk wheat arrives, gets stored and forwarded as per the requirements. In this receival points, the application of traditional quality measurement methods is difficult and often very expensive. Therefore, there is a need for non-invasive, non-destructive real-time methods for wheat quality assessments. One such method that fulfils the above-mentioned criteria is hyperspectral imaging (HSI) for food quality measurement and it can also be applied to bulk samples. In this paper, we have investigated how HSI has been used in the literature for assessing stored wheat quality. So that the required information to implement real-time digital quality assessment methods at the different stages of Australian supply chain can be made available in a single and compact document.
	\end{abstract}


\section{Introduction}
Wheat is the most produced crop in Australia. On average 25 million tonnes of wheat is produced in Australia per annum and it is one of the most important commodity in the global trade scenario \cite{ref1}. Approximately 75\% of the production can be exported to overseas and the remaining is used for domestic market \cite{ref2}.

There are various wheat classes in Australia based on the quality parameters and the suitability for end-use products: Australian Prime Hard (APH), Australian Hard (AH), Australian Premium White (APW), Australian Noodle Wheat (ANW), Australian Standard White (ASW), Australian Premium Durum (ADR), and Australian Soft (ASFT) \cite{ref3}. Each class may have different grades which is allocated by Grain Trade Australia (GTA) based on the physical grain quality. Table \ref{GC_TB} shows how different classes and grades of Australian wheat differs in terms of quality. When a consignment reaches to a receival point in a supply chain, sample of grains are taken out for assessing the physical quality such that, 3 litres of samples are drawn for deliveries up to 10 tonnes with an additional litre for each additional 1o tonnes. Assessment of wheat samples consist of testing for moisture, protein, test weight, screenings,  un-millable material, infected grains, foreign seed contaminants and other contaminants \cite{ref3}. 

Traditional methods of testing wheat quality and safety parameters consist of chemical, biological analysis and human visual inspection, respectively. However, these methods are invasive, destructive, slow and expensive.  
Therefore, the testing paradigm is shifted more towards the faster, non-invasive, non-destructive and real-time/online approaches. 

One of most popular non-invasive and digital approach for quality and safety assessments in agriculture and food industry is hyperspectral imaging (HSI). At the beginning, HSI was used as a tool in remote sensing applications but gradually it has been successfully applied in different applications of agriculture and food industry. Some examples are:  food quality and safety analysis \cite{liu2017hyperspectral,wu2013advanced}, detection of damages in fruits and vegetables \cite{ariana2006near, zhu2019rapid}, detection of contaminants in food products \cite{he2015hyperspectral}, quality analysis of dairy \cite{calvini2020exploring} and meat \cite{feng2020colour} products. HSI has also been widely used cereal grain quality and safety analysis.  
In the literature, there are many research and survey articles can be found on the usage of HSI in food quality and safety analysis including quality assessments of wheat and other cereal grains. However, the literature still lacks a thorough analysis how HSI can be used to assess stored wheat quality at the different stages of supply chain in the Australian context. Therefore, we have written this survey paper where we have discussed how HSI can be used to assess  post-harvest stored wheat quality as per GTA standards \cite{ref3} provided in Table \ref{GC_TB}.


The rest of paper is organised as follows. Section II provides an introduction to HSI. Section III discusses the applications of HSI for wheat protein prediction followed by wheat moisture analysis discussion in Section IV. Detection of different types of defects in wheat grain using HSI is discussed in Section V. Section VI illustrates the detection of contaminants in stored wheat using HSI. Finally, Section VII concludes the paper.

\section{Introduction to Hyperspectral Imaging}

RGB colour imaging is a non invasive and non destructive approach. It deals with three spectral measurements (red, green and blue) within the wavelength range (400—700 nm) \cite{harris2000light} of electromagnetic spectrum to match with the trichromatic nature of the human eyes. 
Different types of features like morphological (shape and size), colour, texture can be extracted from RGB images and they have been successfully used in various applications of agriculture industry. For example, classification of cereal grains using morphological, colour and texture features \cite{majumdar2000classification4, choudhary2008classification}, classification of different grain types and different dockage categories \cite{paliwal2001evaluation, paliwal2003cereal, visen2003image} and identification of damaged kernels in bulk wheat \cite{luo1999identification} using morphological and colour features,  identification of moisture level in wheat kernels using morphological features \cite{tahir2007evaluation, ramalingam2011characterization}. However, RGB image features have some limitations. The effectiveness of these features to analyse the chemical composition of objects is limited. They are also unable to detect the minor constituents and tiny contaminants. In agriculture and food industry, it is required to build a system that can analyse not only the physical properties but the chemical properties as well. In addition, the system should be effective to identify and separate tiny elements. 

In early 2000's, X-ray imaging \cite{karunakaran2003x, neethirajan2007detection} and near-infrared spectroscopy (NIRS) \cite{maghirang2003automated} have shown potential for online, non-invasive and non-destructive applications. However, X-ray imaging present potential health risks as wheat kernels are exposed to radiation and it is an expensive process as well.  NIRS analyse the spectral signatures of the objects and able to detect changes in chemical composition. However, it does not provide any spatial information. Therefore, it is unable to provide any location specific information.

\begin{figure}
	\centering
	\includegraphics[width=1\linewidth]{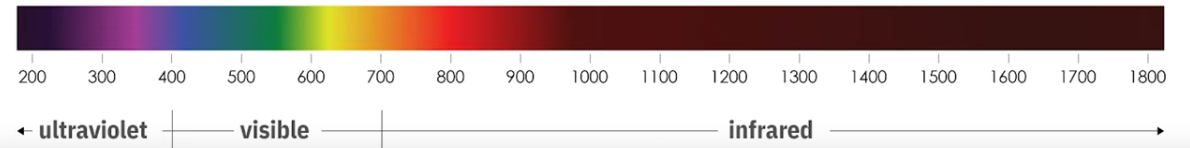}
	\caption{Electromagnetic spectrum wavelength (nm) \cite{em-spectrum}.}
	\label{fig:hsi-wavelength-range1}
\end{figure}

On the other hand, hyperspectral imaging (HSI) combines NIR spectroscopy and digital imaging to provide spatial as well as spectral information of the targets. HSI cameras can operate from the UV range to the near-infrared range and the captured data contains information belonging to a wide range (200 nm- 1300 nm) \cite{sudharsan2019survey} of electromagnetic (EM) spectrum. A visual representation of EM spectrum from ultraviolet to infrared range is given in Figure \ref{fig:hsi-wavelength-range1}.

HSI captured data corresponds to three-dimensional (3D) hypercubes consist of two spatial dimensions and one spectral dimension \cite{gowen2007hyperspectral}. It captures multiple images at different wavelengths for the same spatial area. HSI provides a large amount of data that helps analysing inherent properties of the target. When biological elements are exposed to the EM energy, they reflect, scatter and absorb in a unique pattern at different wavelengths due to their chemical composition and inherent physical structure. This pattern is called spectral signature or spectrum. As it is unique to each element, the spectral signatures can be used to identify or distinguish between different elements \cite{elmasry2012principles}.

Hyperspectral imaging generally measures the intensity of the absorbed or emitted radiation over a range of a spectral band along the EM spectrum. The goal of the HSI is to understand the characteristics of the target object at the molecular level when it is exposed to EM signals. As HSI deals at the molecular level, it can identify the minor changes in the chemical composition of the target objects \cite{srivastava2022detection}.

\begin{figure}
	\centering
	\includegraphics[width=1\linewidth]{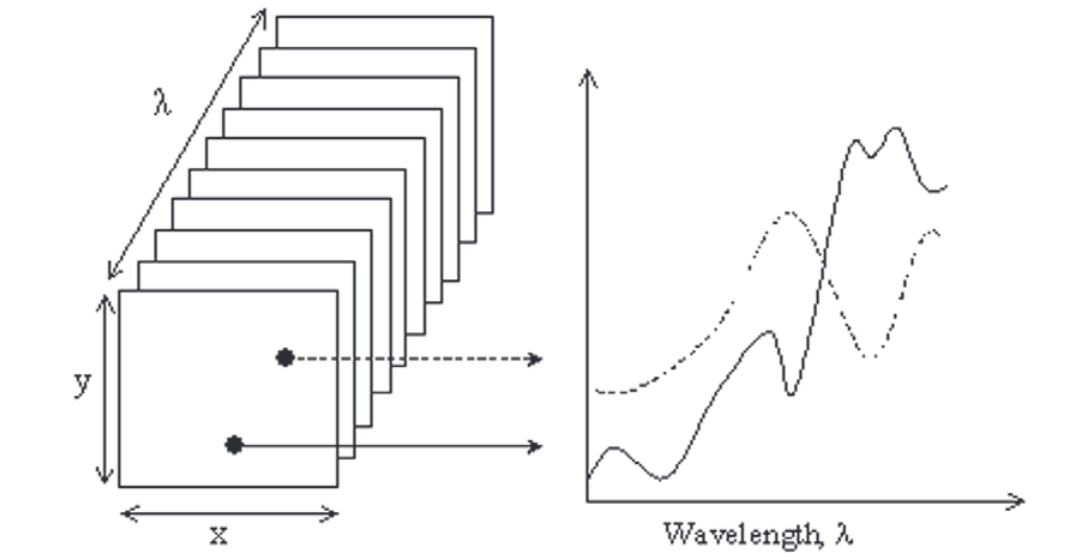}
	\caption{A sample Hypercube showing spatial ($ x $,$ y $) and spectral ($ \lambda $) dimensions (left) and spectra of two separate pixels (right). Adopted from \cite{shahin2008detection}. }
	\label{fig:hypercube}
\end{figure}

Data collected using a hyperspectral imaging system are the absorbed or the emitted radiation values and are stored in the form of a hypercube. The hypercube is a complex data unit, which contains abundant information about the physical and chemical properties of the sample. Hypercubes can be viewed as the stacks of 2D images each captured at a particular wavelength. Each pixel in a hypercube contains the entire spectrum belonging to the imaging bands. These spectra are analysed to obtain the information about the composition of the corresponding pixels. A sample hypercube along with a pair of spectra is provided in Figure \ref{fig:hypercube}.

 After HSI data has been captured, it is pre-processed to extract analytical information and remove scattering effects. At the next step, chemometric analysis to be performed on the pre-processed data for classifying, identifying the elements under consideration or predicting the chemical composition (e.g., protien , moisture, DON level) of elements \cite{del2010early}. Examples of chemometric analysis are principal component analysis, partial least squares, multi-linear regression, linear discriminant analysis, Fischer discriminant analysis and artificial neutral networks \cite{orina2017non}. In this paper, we have reviewed literature belonging to HSI applications for quality assessments of stored wheat. So that an overall idea can be established on how to assess wheat quality at the different stages of supply chain in the Australian context. A list of paper which are reviewed is provided in Table \ref{literature_types}. The detailed discussion is provided in the later sections.

 \begin{table}
 	\centering
 	\caption{List of literature }
 	\label{literature_types}
 	\begin{tabular}{|>{\centering}m{1.2in}|>{\centering\arraybackslash}m{1.8 in} |}
 		\hline
 		Application & Related literature  \\ \hline
 		Protein prediction &   \cite{nir1},\cite{nir2},\cite{nir3},\cite{nir4},\cite{nir5},\cite{nir6},\cite{nir7}, \cite{dowell2006predicting}, \cite{caporaso2018protein}, \cite{engstrom2021predicting}, \cite{he2016deep}, \cite{mahesh2015comparison}, \cite{shuqin2016predicting}, \cite{tahmasbian2021comparison} \\ \hline
 			
 		Moisture prediction &   \cite{wang2004determination}, \cite{dowell2006predicting}, \cite{mahesh2011identification}, \cite{mahesh2014comparing}  \\ \hline
 		Defect detection & 
 		 \cite{shashikumar1993predicting}, \cite{xing2010detection}, \cite{wu2012recognition}, \cite{barbedo2018detection}, \cite{zhang2020non}, \cite{grassi2018monitoring}, 
 		  \cite{berman2007classification}, \cite{delwiche2009hyperspectral}, \cite{shao2020determination}, 
 		   \cite{alisaac2019assessment}, \cite{delwiche2004detection}, \cite{delwiche2008high}, \cite{polder2005detection}, \cite{berman2007classification}, \cite{shahin2011detection}, \cite{weng2021reflectance}, \cite{yipeng2022determination}, \cite{serranti2013development}, \cite{femenias2021standardization}, \cite{nadimi2021examination}, \cite{liang2018determination}, \cite{zhang2020integration}, \cite{zhao2020integration}, \cite{hocking2003microbiological}, \cite{senthilkumar2016detection}, \cite{singh2007fungal}, \cite{liang2020comparison}, \cite{li2022discrimination}
 		\cite{ridgway1998detection}, \cite{singh2009detection}, \cite{singh2010identification}, \cite{murdoch}, \cite{capsulanet}
 		\\ \hline
 		Contaminant detection & \cite{pierna2012nir}, \cite{ravikanth2015classification}, \cite{ravikanth2016performance}, \cite{ravikanth2016detection}, \cite{vermeulen2012online} \\ \hline
 
 	\end{tabular}
 \end{table}

\section{HSI for protein prediction}

Protein is one of the most important wheat quality parameter due to its impact in baked products. Therefore, wheat price is dependent on protein content. Based on this, bonus payments or discount offers are being made on top of the base rate for each grade \cite{taylor2005monitoring}. Before the application of HSI, Near-infrared spectroscopy (NIRS) was used as the digital method to investigate the wheat protein content. Initially, NIRS was possible to be applied only on ground wheat or flour\cite{nir1}. As it is destructive approach, investigations on non-destructive NIRS application on bulk of whole whole wheat were initiated in \cite{nir2,nir3}. Among individual wheat kernels, protein content varies \cite{nir6}. However, analysis of bulk wheat samples, these variability cannot be detected \cite{nir4}. Therefore, NIRS analysis has been applied to individual kernels as well \cite{nir4,nir5}. It has been also shown that protien prediction of bulk wheat can be improved by averaging the individual kernel measurement of the corresponding bulk \cite{nir7}. Partial least squares (PLS) regression was applied for protein prediction on the NIRS data of single wheat kernels in \cite{dowell2006predicting}.

Development of an HSI calibration was proposed in \cite{caporaso2018protein} to measure and visualise protein distribution within single wheat kernels by assessing the uniformity present in the wheat samples.  A deep-learning based regression model has been proposed in \cite{engstrom2021predicting} for protein prediction in bulk grain by processing the entire hypercube (spatial and spectral data) using a modification of ResNet-18 \cite{he2016deep}, a popular convolutional neural network architecture. A comparison of two regression models:  Partial Least Squares Regression (PLSR) and Principal Components Regression (PCR) has been performed for protein prediction using NIR hyperspectral images of Canadian wheat in \cite{mahesh2015comparison} and experimental results show PLSR outperform PCR. Similar observation was obtained in \cite{shuqin2016predicting} to detect protein content using HSI of 11 varieties of wheat. However, in this case, protein prediction accuracy was improved by applying radial basis function (RBF) neural network.

Traditionally, protein content calculation in a wheat sample is dependent on its nitrogen (N) content. As per GTA standards, the conversion factor to obtain the wheat protein from nitrogen is 5.7. i.e., $ Protein = Nitrogen \times 5.7 $. However, in practice, this conversion factor ranges from $ 5.7 - 6.25 $ as different testing laboratories follow their own conversion factors \cite{N-P, mariotti2008converting}. As N is an indicator of wheat protein content, N prediction like in \cite{tahmasbian2021comparison} can be used for calculating protein prediction where 
a SWIR imaging and PLS regression  have been performed to predict the N concentrations in wheat samples and it was found that most important spectral bands are 1451–1600 nm, 1901–2050 nm and 2051–2200 nm.

\section{HSI for moisture prediction}

In Australia, grains are traditionally harvested at a lower moisture content (mostly under 13\%) after it has been sun dried to avoid the extensive drying process post harvest \cite{taylor2005monitoring}. Although post maturity sun drying can decrease the quality and early harvesting with higher moisture can increase the yield \cite{banks1999high}, the lower the moisture content, the higher advantage for buyers as higher moisture wheat are prone to insect infestation, mould growth, reduced germination \cite{nanje2013use}.   

Therefore, precise identification of moisture content in wheat is crucial. Higher level of moisture can cause spoilage and sprouting in wheat kernels. At a supply chain receival point, wheat samples may arrive from different sources of variable moisture levels and kept in storage together prior to further processing. Therefore, before mixing the wheat from various sources, it is necessary to dry high-moisture wheats to an optimum level to prevent spoiling the entire storage bin.

Similar to protein analysis, NIRS was also used for moisture prediction in wheat as non-invasive digital method. Initially NIRS was used for moisture prediction in ground wheat \cite{wang2004determination} but later on it was also used for whole wheat kernels in \cite{dowell2006predicting} where PLS regression was performed on NIRS wheat data for moisture content analysis after removing the influence of the protein content to avoid the correlation between moisture and protein. 

HSI was used in \cite{mahesh2011identification} to classify five different moisture levels in western Canadian wheat classes in the wavelength range of 960-1,700 nm. Also each wheat class is identified at five different moisture levels (12, 14, 16, 18 and 20\%) at four effective wavelengths of 1,060, 1,090, 1,340, and 1,450 nm.   
This work has been extended to classify moisture-specific wheat classes in \cite{mahesh2014comparing}.



\section{HSI for Grain defect detection}

The percentage of defective kernels in a bulk of wheat is a determining factor to classify wheat in to the major milling grades. In Australia, the maximum allowable percentage of defective grains due to different factors is standardised by Grain Trade Australia (GTA) \cite{ref3}. Traditionally, defects were detected by visual inspection by examining the kernels. However, these methods are time consuming, labour intensive and expensive. Moreover, often visual identification of defective grains are challenging as they may look similar to the sound kernels.

\subsection{Sprouting}

\begin{figure}[h]
	\centering
	\includegraphics[width=0.9\linewidth]{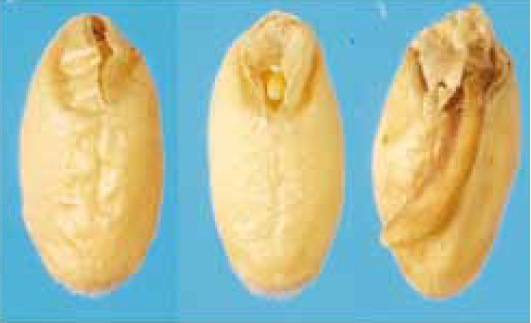}
	\caption{Different sprouting stages in a wheat kernel. Adopted from \cite{ref4}.}
	\label{fig:sprout-grains}
\end{figure}

One of the main cause for grain defect is sprouting.  The baking quality of flour produced from sprouted wheat is poor. Therefore, sprouting is one of the main determining factor for assessing wheat in different grades. Sprouting occurs as a result of germination due to rain damage by absorbing water at the pre-harvest stage. Rain damage causes the development of the $\alpha$-amylase and other germinative enzymes which negatively impact the flour produced from the sprouted wheat. In addition, sprouted wheat kernels are prone to infestation by insects \cite{ref4}.  In Australia, there is zero tolerance of sprouted wheat for human consumption \cite{ref3}. A sample image showing different stages of sprouting in wheat is provided in Figure \ref{fig:sprout-grains}.

Traditional pre-harvest rain damage detection methods like Hagberg Falling Number test \cite{ref4} is destructive and time-consuming. Therefore, non-destructive methods were started to be explored. X-rays used to classify sprouted and sound wheat kernels in \cite{neethirajan2007detection} but X-rays present potential health risks as wheat kernels are exposed to radiation.  

Similar to other parameters of assessing wheat quality, HSI also has been explored to detect rain-damage in wheat kernels. It was demonstrated in \cite{shashikumar1993predicting} that NIR spectra can be used for damage detection due to sprouting but, the effectiveness was compromised as only three wavelengths were considered in 1445–2345 nm range of NIR spectra. An equally spaced 60 wavelengths in the range 1000-1600 nm has been used for NIR HSI for classifying sprouted and healthy wheat kernels. From the results, three specific wavelengths 1101.7, 1132.2, and 1305.1 nm were identified as significant for the analysis. A visible near-infrared (VNIR) HSI in the range 400-1000 nm was employed to classify sprouted, severely sprouted and healthy kernels. Average spectra of sprouted and severely sprouted kernels show higher reflectance responses compared to the healthy kernels in the wavelength region above 720 nm. Principle component analysis (PCA) was used to pick four suitable wavelengths. It has also been shown that by taking the ratio of reflectance between 878 nm and 728 nm, healthy kernels can be distinguished from the sprouted ones.\cite{xing2010detection}. 

In \cite{wu2012recognition}, authors claimed that when HSI analysis performed at VNIR range, at 675 nm,
a significant difference for spectral reflectivity between sprouting and healthy kernels was noticed. It was also shown
that the imaging spectra can differentiate different levels of wheat preharvest sprouting. To discriminate between sprouted and sound kernels using NIR HSI, it was observed that, there is a visible difference between them in two spectral bands which are 844-1140 nm and 1386-1700 nm, respectively. In the first band (844-1140 nm), sound kernels have a higher reflectance 10-20\% than the sprouted ones. In contrast, the reflectance of sprouted kernels is 20\%-60\% larger than the sound kernels in second band(1386-1700 nm) \cite{barbedo2018detection}. It is difficult to identify slightly sprouted kernels by visual inspection and it can deteriorate the quality of sound kernels if mixed together. Therefore, in \cite{zhang2020non} NIR HSI has been used in 866.4 nm - 1701 nm wavelength range to differentiate between slightly sprouted and sound kernels. The authors have also investigated the effect of imaging of each kernel from both sides and found that the result is better when considering reverse (dorsal) side than the front (ventral) side. Wheat kernels undergo changes in chemical composition, enzymatic activities, starch and protein contents when affected by rain damage. A portable NIR device which works in the wavelength range of 950–1650 nm has been employed to monitor these changes during the germination process to determine the level of sprouting \cite{grassi2018monitoring}.

\subsection{Black point}

\begin{figure}[h]
	\centering

	\includegraphics[width=0.9\linewidth]{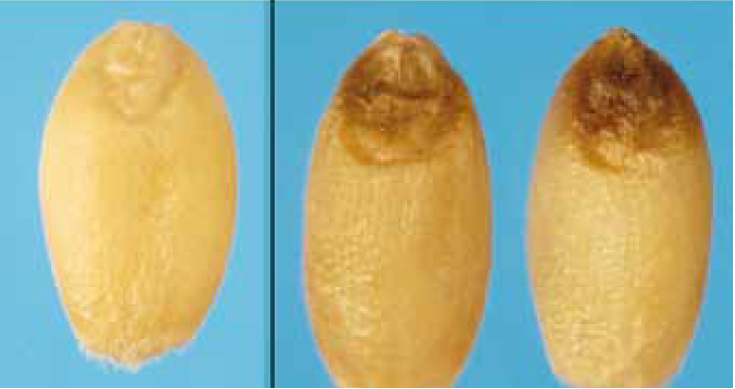}
	\caption{(left): Sound kernel; (right): Black point affected kernel. Adopted from \cite{ref4}.}
	\label{fig:black-point}
\end{figure}

Another major grain defect is black point or black tip that begins in the germ and extends around one or both cheeks and into the crease. It is caused by moisture and damp conditions during the ripening of wheat. Black point affected wheat considered to be not fit for human consumption and therefore, needs to be removed \cite{ref4}. A sample image showing black point affected kernels is provided in \ref{fig:black-point}.

Classification of black point affected kernels and sound kernels has been conducted in \cite{berman2007classification} using HSI. The authors have shown that the results have minimal effect if shorter wavelength ranges ( 420-1000 nm or 420-700 nm) have been used instead larger one (420-2500 nm). Therefore, higher classification accuracies can be achieved considering a few wavelengths and the process can be made efficient and less-expensive. In \cite{delwiche2009hyperspectral}, a single fluorescence wavelength (531 nm) was selected for spectral image processing to differentiate between sound and black point-damaged kernels. Authors have concluded that the effectiveness of detection of black points are higher when wheat kernels are imaged from the dorsal side. This is because, the damaged region is comparatively smaller in the ventral side. Also the crease in the ventral side may create a shadowing effect which can cause confusion with the black point. 

Individual damage or defect detection in wheat kernels may be time consuming and expensive where a bulk of wheat samples are needed to be analysed faster. Therefore, in \cite{shao2020determination}, an HSI approach in 865-1711 nm wavelength range is proposed to identify damaged (including black point) kernels from a bulk of wheat. At first, effective wavelengths were obtained separately by the application of PCA and successive projection algorithm (SPA). After that, Partial least square (PLS) and least square-support vector machine (LS-SVM) classifiers were built using both full spectral and effective wavelength data. It has been observed that LS-SVM outperforms PLS and results with full spectral and effective wavelength data were comparable. 

\subsection{Fungal infection}

\begin{figure}[h]
	\centering
	\includegraphics[width=0.9\linewidth]{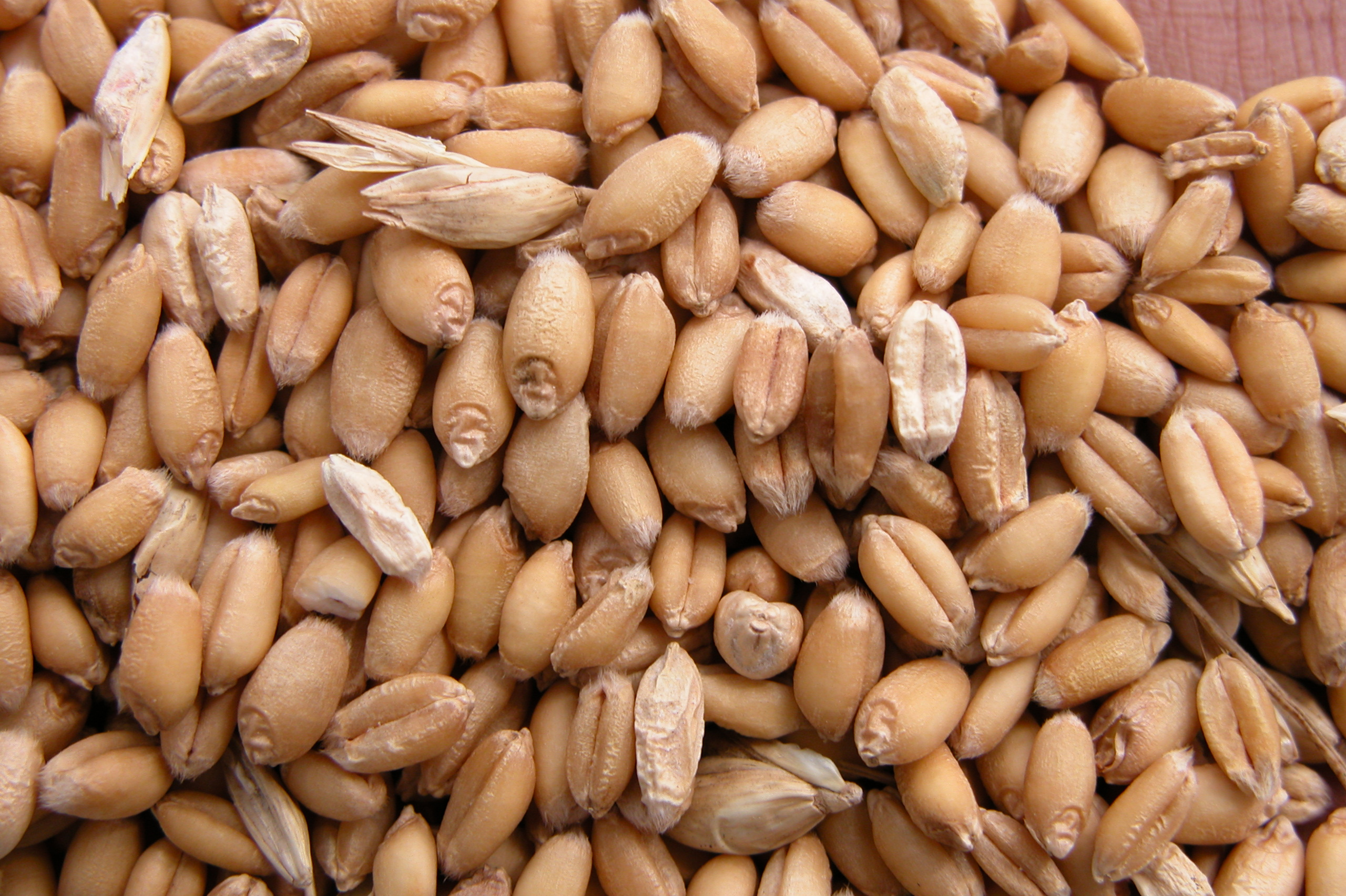}
	\caption{A sample of wheat containing sound and FHB damaged kernels. FHB damaged kernels are shrivelled and pale compared to the healthy ones. Adopted from \cite{fhb-wa}.}
	\label{fig:fhb}
\end{figure}

Fungal infection causes severe damage in wheat kernels. Fusarium head blight (FHB) or fusarium damage is another major defect in wheat kernels caused mainly by a fungus name Fusarium graminearum \cite{goswami2004heading}. Other species such as Fusarium culmorum, and Fusarium poae can also cause FHB damage 
\cite{alisaac2019assessment}. Fusarium infection causes the production of mycotoxin deoxynivalenol (DON), a toxic for humans and livestocks. Also, it reduces the kernel weight, affects flour colour and baking performance. Therefore, infected kernels need to be removed. A sample containing sound and FHB damaged kernels is provided in Figure \ref{fig:fhb}.

Traditional approach of detecting fungal infection by visual inspectors is not suitable for rapid online testing. Thus, effective digital solutions are required. NIR spectroscopy has been successfully used to separate Fusarium-infected and sound kernels under controlled laboratory conditions \cite{delwiche2004detection} but under commercial operating conditions, the NIR spectroscopy performance is not satisfactory \cite{delwiche2008high}. Imaging in short-wave infrared range (1000-2500 nm) has also been used FHB detection \cite{polder2005detection} but the cameras operating in the short-wave infrared range are very expensive and cannot be afford by many. To overcome this situation, it has been shown in \cite{berman2007classification} that the HSI using 420–1000 nm wavelength range provides comparable fungal-infection detection. 

\begin{figure}[h]
	\centering
	\includegraphics[width=0.9\linewidth]{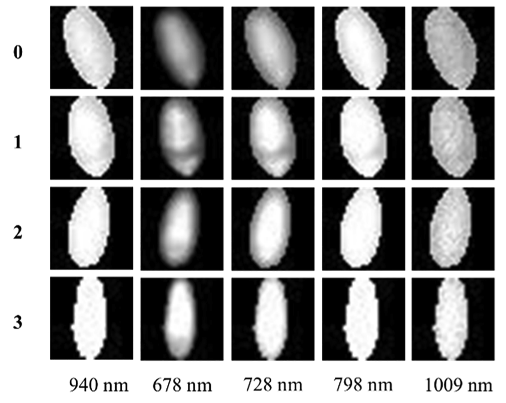}
	\caption{Reflectance images at five EWs of sound and FHB damaged kernels. Class 0 to healthy and 1-3 corresponds to mild, moderate and severe damaged kernels. Adopted from \cite{weng2021reflectance}.   }
	\label{fig:fhb-severity}
\end{figure}

In \cite{shahin2011detection}, mildly and severely fusarium-damaged wheat kernels were distinguished from the sound wheat kernels using HSI in the VNIR wavelength range. Spectral images pre-processed by PCA to reduce the data to 10 scores upon which linear discriminant analysis (LDA) models were developed to classify sound and fusarium-damaged kernels. The damaged kernels were further classified into mild and severe. It has also been shown that optimally selected six wavelengths (484 nm, 567 nm, 684 nm, 817 nm, 900 nm and 950 nm) can exhibit comparable results that with the full VNIR spectrum. Different degrees of FHB-damage (e.g., mild, moderate, severe) in wheat kernels is identified in \cite{weng2021reflectance} using HSI and deep neural networks. At first, five effective wavelengths (EW) were selected from preprocessed raw HSI data. After that the combination of reflectance images (RI) corresponding to the EWs as seen in Figure \ref{fig:fhb-severity} were analysed using LeNet-5 and residual
attention convolution neural network (RACNN) to identify the degrees of FHB-damage in wheat kernels. It was concluded that the optimal performance was achieved for the combination of RIs at 940 nm and 678 nm.

Another work on identifying the degree of FHB-damage in wheat kernels using HSI and deep neural networks is proposed in \cite{yipeng2022determination}. The deep learning network is neural architecture search (NAS)-based. HSI has been performed in 400-1000 nm range. Effective wavelengths (EWs) were selected for reflectance spectroscopy of HSI and corresponding monochromatic images were combined with an architecture self-search deep network (ASSDN) to develop the self generated and self-optimised determination model for the degree of FHB damage.

An HSI approach in NIR range (1000-1700 nm) has been considered in \cite{serranti2013development} to separate fusarium-damaged kernels from the mixture of yellow berry kernels (acceptable amount) and vitreous (sound) kernels. In addition, yellow berry kernels were also separated from vitreous kernels when top quality wheat is required. After compressing the data using PCA, partial least squares discriminant analysis (PLS-DA) was applied to classify vitreous, yellow berry and fusarium-damaged kernels. In addition,  interval PLS-DA (iPLS-DA) was used for the same classification in three shorter wavelength ranges. It has been shown that classification results obtained using the entire wavelength range and the shorter wavelength ranges are comparable.  
NIR HSI method has also been used in \cite{femenias2021standardization} for processing single wheat kernel to classify them according to fusarium damage, followed by determination of DON level in infected kernels and finally classify infected kernels based on DON level. For HSI, single wheat kernels were scanned both crease-up and crease-down position. After spectral preprocessing the raw hyperspectral images, the data corresponding to the optimally selected wavelengths used to train and validate machine learning models (LDA, Naive Bayes, KNN) to achieve the classification and regression predictions. Another NIR HSI method in the 900-1700 nm wavelength range has been proposed for detection of fusarium damage and DON content in \cite{nadimi2021examination}.

 An online rapid detection of FHB-infected wheat kernels followed by the determination of DON level in the infected kernels using HSI was proposed in \cite{liang2018determination}. Imaging has been performed in 400–1000 nm. Two spectral preprocessing methods: Standard normal variate (SNV) transformation and
 multiplicative scatter correction (MSC) combined independently with two wavelength selection methods: successive projections algorithm
 (SPA) and random frog (RF), generated four different combinations of optical wavelength. After the pre-processing, optimally selected data were used to build classification models using SVM and PLS-DA classifiers. The combination of MSC–SPA–SVM resulted in the highest classification accuracy of 97.92\% on the test set. After that, the DON levels in the FHB-infected kernels were also determined from the predicted values of classifiers.

In \cite{zhang2020integration}, grayscale image features were fused with spectral data to identify the degree of FHB-damage in wheat kernels. After pre-processing the raw HSI data, optimal wavelengths were selected by applying successive projections algorithm with random forest (SPA-RF) method. Image features were extracted from the images corresponding to the optical wavelengths and combined with spectral features. This fused features were then used to build classification models to identify the sound or different degrees of damaged wheat kernels. During the analysis of reflectance spectra of HSI data, it was noticed in \cite{zhao2020integration} that the more severe the FHB infection is, the greater the reflection is. This is because the toxin released by the FHB destroys the cell wall of the wheat to form a hole like
structure between the seed coat and aleurone layer. Therefore, it causes the increase in reflectance.

Formation of mycotoxins by fusarium in post harvest dry stored grain is limited. In contrast, damage in stored grains are more prone to fungi like Aspergillus and Penicillium \cite{hocking2003microbiological} and produces a toxin named Ochratoxin A. The damage due to these fungi and the resulting toxins were detected using an NIR HSI approach in \cite{senthilkumar2016detection}. After preprocessing the HSI data, six statistical and ten histogram features were extracted from the data corresponding to the selected effective wavelengths. The extracted features were then used for developing linear, quadratic and Mahalanobis discriminant classifiers. It has been found that wavelengths 1280, 1300, and 1350 nm were crucial to detect fungal damage and to detect Ochratoxin A. 
Among the classifiers, quadratic discriminant classifier outperform the other two. Similar work of detecting the damage due to Aspergillus and Penicillium was proposed in \cite{singh2007fungal} where HSI in 1000-1600 nm range was performed. After that the PCA-applied dimensionality-reduced spectral data was used to build linear and quadratic discriminant classifiers to distinguish between damaged and sound kernels. 

A comparison of VNIR and SWNIR imaging have been performed for detecting DON levels in fusarium damaged wheat kernels in \cite{liang2020comparison}. The changes occurred in the protein, carbohydrate and lipid contents of the damaged kernels due to mycotoxins (DON) cannot be analysed visually. Therefore, to investigate the effectiveness of damage detection by using the spectral characteristics, this comparison has been performed. Data captured at the VNIR and SWNIR wavelength ranges were preprocessed by either multiplicative scatter correction (MSC) or standard normal variate (SNV) followed by the application of genetic
algorithm (GA) to select the effective wavelengths. Support vector machine (SVM) and sparse autoencoder(SAE) network were used to build the classifiers and it was found that the detection of DON in damaged kernels is more effective in VNIR rather than SWNIR. Although the performances of SVM and SAE were comparable, it was found that the combination of MSC-GA-SAE produced the highest accuracy.

In practical experiment scenario, it often happens that HSI samples available to build machine learning models to identify sound and fungal-infected wheat kernels is limited. Especially, the samples of damaged/infected kernels are found lesser compared to the samples of sound kernels. This data imbalance can cause compromising results generated from machine learning models. To overcome this issue, in \cite{li2022discrimination} deep convolutional generative adversarial network (DCGAN)-based data augmentation has been used to increase the training data from the raw NIR HSI data. Experimental results show that by expanding the training samples, classification accuracies of different machine learning models has been increased significantly.

\subsection{Insect-damage}

\begin{figure}
	\centering
	\includegraphics[width=0.16\textwidth, angle =90]{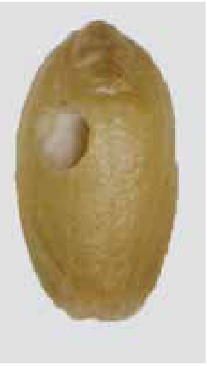}
	\caption{Insect-damaged wheat kernel. Adopted from \cite{insect-damage}. }
	\label{fig:insect-damaged-wheat}
\end{figure}

Damage caused by insects such as rice weevil (Sitophilus oryzae), lesser grain borer (Rhyzopertha dominica) \cite{wheat-insects} is a major concern in wheat storage locations. Wheat grains result in weight and nutrient loss, reduced germination ability and prone to fungal infection. Therefore, quality of the wheat grains deteriorate significantly and causes reduced market value. In addition, presence of insect fragments degrade the milling quality \cite{perez2003detection}. An example of insect-damaged wheat kernel is shown in Figure \ref{fig:insect-damaged-wheat}.

\begin{figure}
	\centering
	
	\subfigure[]{\includegraphics[width=0.24\textwidth]{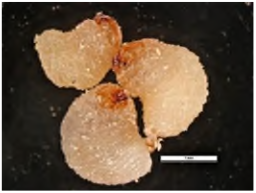}}  
	\subfigure[]{\includegraphics[width=0.24\textwidth]{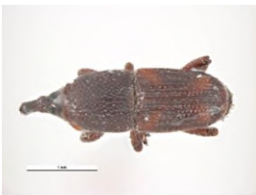}} 
	
	\caption{(a) Larvae and (b) Adult stages of Rice weevil (Sitophilus oryzae). Adopted from \cite{insects}.} 
	\label{fig:insect}
\end{figure}

 Mainly two types of insects damage wheat kernels, they are primary internal feeders and secondary feeders. Former types are difficult to detect, they develop inside the wheat kernels and get out after growing up, leaving holes in the kernels. At the start of life cycle, female insects create a tiny hole in the healthy kernels and deposit eggs there. The kernel cavity is covered by a gelatinous secretion. The eggs hatch inside the kernel, feed and the grown up insects emerges out of the wheat kernels. Often kernels infested by primary feeders contain immature stages of insects (eggs, larvae, and pupae) but they look apparently same as sound kernels. Later types survive on already damaged and broken kernels \cite{pearson2003automated}. A sample image of larvae and adult rice weevil is shown in Figure \ref{fig:insect}.

Respiratory and metabolic activities of insects generate heat and moisture which can cause local hotspots in the grain storage and can result in fungal infection \cite{manickavasagan2006thermal, local-hotspot}. Traditional methods can only detect visual damages or traces of insects but unable to detect internal insect infestation. HSI can address this issue even with the spatial location of the infestation. When wheat kernels are infested, there are significant changes occur in their moisture and starch content \cite{el2006detection}. These changes can be easily detected by HSI unlike the traditional methods. 

\begin{figure}
	\centering
	\includegraphics[width=.8\linewidth]{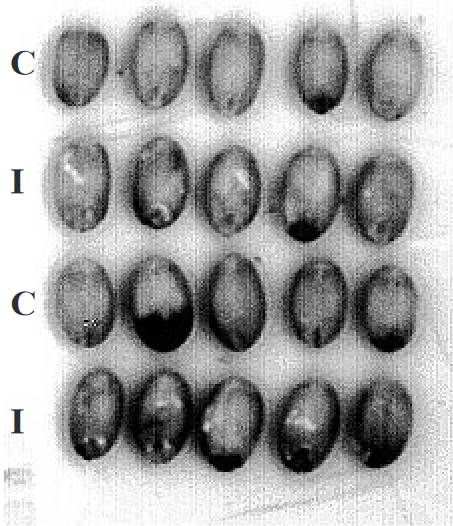}
	\caption{C: sound kernels; I: internally infested kernels imaged at 1202 nm and 1300 nm, respectively. Adopted from \cite{ridgway1998detection}. }
	\label{fig:insect-damage}
\end{figure}

In \cite{ridgway1998detection}, it has been found that images of internally infested and sound wheat kernels captured at two different wavelengths (1202 and 1300 nm) show evident differences. Internally infested kernels appear brighter. Whereas, the sound kernels appear darker as can be seen in Figure \ref{fig:insect-damage}. Although HSI analysis in \cite{ridgway1998detection} effective to distinguish insect-damaged kernels, a machine learning model build with training data is necessary to process unknown samples. Therefore, supervised classification models using long-wave NIR HSI has been developed in \cite{singh2009detection} to classify sound and infested kernels using the training samples as images captured in the range of 1000-1600 nm. After preprocessing and dimensionality reduction using multivariate analysis, two wavelengths (1101.69 and 1305.05 nm) were shortlisted. Six statistical features and 10 histogram features from the images corresponding to the shortlisted wavelengths were extracted to build statistical discriminant classifiers. As working in the long-wave NIR region is  expensive, the same authors have investigated to build the classifiers in short-wave NIR range (700 - 1100 nm) in \cite{singh2010identification}.

Researchers from Murdoch University, Australia have developed a system name Capsulanet using VNIR HSI to identify different types of insects in the stored grain setting \cite{murdoch}. More than 2000 images of adults, larvae and larvae skins of different types of insects have been captured to analyse the reflected energies at different wavelengths. This analysis helps to determine the physical and chemical characteristics of the insects and eventually detecting them \cite{capsulanet}.

\section{HSI for Contaminants detection}

\begin{figure}
	\centering
	
	\subfigure[]{\includegraphics[width=0.14\textwidth]{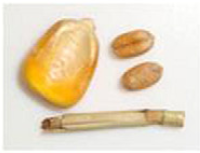}}  
	\subfigure[]{\includegraphics[width=0.12\textwidth]{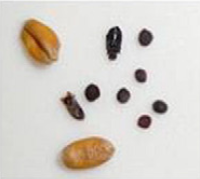}} 
	\subfigure[]{\includegraphics[width=0.17\textwidth]{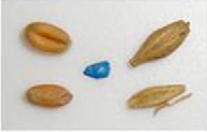}} 
	
	\caption{Images of (a) two wheat kernel, a straw and one maize kernel (b) two wheat kernels, six rapeseeds and two insects (c) two wheat kernels, two chaffy barley kernels and a piece of painting. Adopted from \cite{pierna2012nir}. } 
	\label{fig:cont}
\end{figure}

The presence of unwanted materials is a major determining factor for assessing wheat quality. This contaminants include 
un-millable materials such as, broken, small and cracked kernels. In addition, foreign seeds and other contaminants are also unwanted. Examples of foreign seed contaminants are poppy, coriander, peanut, soybean, sunflower, chickpea, broad bean etc. As per GTA standard \cite{ref3}, they fall into 11 categories. Examples of other contaminants are ryegrass ergot, insects, snails, loose smut, sand etc. Some contaminants are often lighter than the sound wheat kernels and contain higher moisture. A sample of various types of contaminants coexist with wheat is shown in Figure \ref{fig:cont}. Bulk wheat are stored in silos and the lighter contaminants gather along the wall of silos. This makes the high-moisture, lighter contaminants more prone towards mould growth and eventually damage the wheat of the entire silo. Therefore, accurate identification followed by removal of these unwanted materials is necessary for long-term safe storage.

In addition, any appalling item like animal excreta contributed by the the living animals like insects, snails, rodents contaminate wheat is considered as filth \cite{dogan2017analysis}. Visual presence of filth reduces the wheat quality and its marketability. Therefore, it is required to identify and remove these animals from the bulk wheat storage.

Traditional methods like visual inspection to detect contaminants has many drawbacks and therefore, to facilitate fast and online detection, HSI has also been used for contaminant detection. In \cite{pierna2012nir}, NIR HSI in the wavelength range of 1100-2400 nm has been used to detect insects, irrelevant cereal grains, botanical impurities in the bulk grain. Another NIR HSI in the wavelength range of 1000-1600 nm has been used to classify seven foreign material types (barley, canola, maize, flaxseed, oats, rye, and soybean), six dockage types
(broken wheat kernels, buckwheat, chaff, wheat spikelets, stones, and wild oats), and two animal excreta types (deer and rabbit droppings). The raw NIR spectra were pre-processed by five different spectral processing methods followed by classifying the pre-processed data by three classification models. Each classification model has been used in two approaches. First one is two-class classification: classification of each contaminant type from the sound wheat kernels. Another approach is multi-class classification: classification of all contaminant types from sound wheat kernel \cite{ravikanth2015classification}. Among the different combinations,  highest classification accuracies were observed when spectral data were pre-processed using the SNV
technique and classified using the k-NN classifier \cite{ravikanth2016performance}.

Two regression models: principal component regression (PCR) and partial least-squares regression (PLSR) were developed to predict the percentage level (0, 3, 6, 9, 12, and 15\%) of broken wheat kernels. Hyper spectral images in 1000–1600 nm wavelength range at a
10 nm interval of bulk wheat samples at different broken level were captured and the corresponding reflectance spectra were used to train the regression models. In addition, classification models were developed using linear discriminant analysis (LDA) and quadratic discriminant analysis (QDA) to successfully classify sound kernels from different levels of broken kernels in bulk wheat samples \cite{ravikanth2016detection}.

\begin{figure}
	\centering
	\includegraphics[width=1\linewidth]{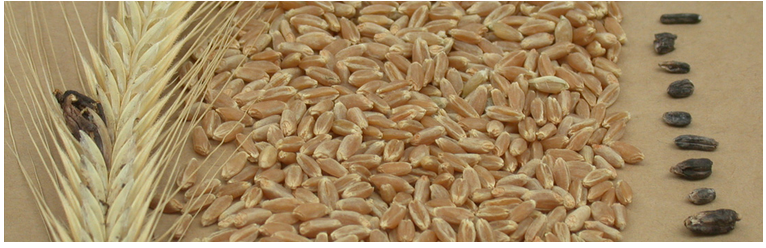}
	\caption{Sound wheat kernels (left) are compared to ergot bodies (right). Adopted from \cite{ergot}.}
	\label{fig:ergot-bodies}
\end{figure}

Cereal grains like wheat are often affected by fungal infection. One such fugal species is Claviceps purpurea, it produces ergot bodies. Ergot bodies possess alkaloid content and it is toxic for both human and animals \cite{agriopoulou2021ergot}. In Australia, some of the major contaminants are ryegrass ergot and cereal ergot \cite{blaney2009alkaloids}. The toxicity of ergot bodies are followed in the wheat flour as well as in the baked products. Therefore, these bodies needs to be identified and removed. Often these bodies are almost the same size and shape as a wheat kernel and replace healthy wheat kernels \cite{ergot}. It makes the ergot removal even difficult. A sample images in Figure \ref{fig:ergot-bodies} shows the visual comparison of sound wheat kernels and ergot bodies.
Different countries have their tolerance levels for ergot bodies. In Australia, the maximum allowable ergot bodies in the major milling grades is assigned by GTA \cite{ref3}.

An NIR HSI method has been proposed in \cite{vermeulen2012online} to detect and quantify ergot bodies. Images were collected using an NIR
hyperspectral line scan combined with a conveyor belt at 209 wavelengths in the range of 1100–2400 nm. For training and evaluation of regression models (e.g., partial least square discriminant analysis, PLSDA), SVM, several samples of wheat with different ergot contamination level were prepared. Evaluated results exhibited a correlation higher than 0.99 between the predicted values and the ground-truth values.

\section{conclusion}

In this paper we have reviewed literature related to post-harvest wheat quality assessment. In a supply chain, wheat trucks arrives as well as leaves the receival points where wheat from different sources may get mixed and stored for a limited time. Therefore, it is required to constantly assess the wheat in this receival points to avoid any degradation or damage to the wheat quality. Traditional assessment methods are invasive, destructive, time-consuming and often expensive. Therefore, non-invasive, non-destructive, real-time or digital methods gained popularity as efficient and effective alternatives. Wheat quality assessment using HSI is one of the most popular alternative. In this  paper, we have discussed how different wheat quality parameters from protein and moisture prediction to defect and contaminant detection have been performed in the literature using HSI. Therefore, this article can be regarded as a reference while implementing digital methods for assessing wheat quality at the different stages of Australian supply chain.

\bibliographystyle{IEEEtran}
\bibliography{grainQuality.bib}

	\begin{landscape}
	\begin{table}
		\caption{Wheat quality standards from GTA \cite{ref3} } \label{GC_TB} 
		\begin{center}
			\small
			\begin{supertabular}{|c|c|c|c|c|c|c|c|c|c|c|c|c|}

				\hline
				\textbf{Class} & \multicolumn{2}{c|}{\textbf{\thead{Australian\\Prime\\Hard\\(APH)}}} & \multicolumn{2}{c|}{\textbf{\thead{Australian\\Hard\\(AH)}}}& \multicolumn{2}{c|}{\textbf{\thead{Australian\\Premium\\White\\(APW)}}} & \textbf{\thead{Australian\\Standard\\White\\(ASW)}} & \textbf{\thead{Australian\\Noodle\\Wheat\\(ANW)}} & \textbf{\thead{Australian\\Soft\\(ASFT)}} & \textbf{\thead{Australian\\Durum\\(ADR)}} & \textbf{\thead{Australian\\General\\Purpose}} & \textbf{\thead{Australian\\Feed\\ }} \\ \hline 
				\textbf{\thead{\\ Grade \\ \\}  } & \textbf{APH1} & \textbf{APH2} & \textbf{H1} & \textbf{H2} & \textbf{APW1} & \textbf{APW2} & \textbf{ASW1} & \textbf{ANW1} & \textbf{SFTl} & \textbf{ADR1} & \textbf{AGP1} & \textbf{FED1} \\ \hline 
				\textbf{\thead{\\ Moisture Max (\%)\\ \\}} & 12.5 & 12.5 & 12.5 & 12.5 & 12.5 & 12.5 & 12.5 & 12.5 & 12.5 & 12.5 & 12.5 & 12.5 \\ \hline 
				\textbf{\thead{\\ Protein Min (\%) (N x  5.7) \\ @ 11\% Moisture Basis \\ \\}} & 14.0 & 13.0 & 13.0 & 11.5 & 10.5 & 10  & N/A & 9.5 & N/A & 13.0 & N/A & N/A \\ \hline 
				\textbf{\thead{\\ Test Weight \\ Min (kg/hl) \\ \\}} & 76.0 & 76.0 & 76.0 & 76.0 & 76.0 & 76.0 & 76.0 & 76.0 & 76.0 & 76.0 & 68.0 & 62.0 \\ \hline 
				\textbf{\thead{\\ Screenings Max \\ (\% by weight) \\ \\}} & 5.0 & 5.0 & 5.0 & 5.0 & 5.0 & 5.0 & 5.0 & 5.0 & 5.0 & 5.0 & 10.0 & 15.0 \\ \hline 
				\textbf{\thead{\\ Unmillable Material \\ Max (\% by weight \\ \\)}} & 0.6 & 0.6 & 0.6 & 0.6 & 0.6 & 0.6 & 0.6 & 0.6 & 0.6 & 0.6 & 1.2 & 2.6 \\ \hline 
				\textbf{\thead{\\ Falling Number\\Min (seconds) \\ \\}} & 350 & 350 & 300 & 300 & 300 & 300 & 300 & 300 & 300 & 300 & 200 & N/A \\ \hline 

			\end{supertabular}
			
		\end{center}
	\end{table}
\end{landscape}

\end{document}